\documentclass[10pt,twocolumn,letterpaper]{article}
\usepackage{iccv}

\usepackage{algorithm}
\usepackage{algpseudocode}
\usepackage{placeins} 
\usepackage{todonotes}
\usepackage{adjustbox}
\usepackage{graphicx}
\usepackage{balance}

\definecolor{iccvblue}{rgb}{0.21,0.49,0.74}
\usepackage[pagebackref,breaklinks,colorlinks,allcolors=iccvblue]{hyperref}

\def\paperID{8} 
\def\confName{ICCV}
\def\confYear{2025}

\title{Now You See Me, Now You Don’t: A Unified Framework for Expression Consistent Anonymization in Talking Head Videos}

\author{Anil Egin\\
Inria Center at Université Côte d'Azur\\
Sophia Antipolis, France\\
{\tt\small anil.egin@studbocconi.it}
\and
Andrea Tangherloni\\
Bocconi University\\
Milan/Italy\\
{\tt\small andrea.tangherloni@unibocconi.it}
\and
Antitza Dantcheva\\
Inria Center at Université Côte d'Azur\\
Sophia Antipolis, France\\
{\tt\small antitza.dantcheva@inria.fr}
}

\begin{document}

\maketitle

\begin{abstract}

Face video anonymization is aimed at privacy preservation while allowing for the analysis of videos in a number of computer vision downstream tasks such as expression recognition, people tracking, and action recognition.
We propose here a novel unified framework referred to as AnonNET, streamlined to de-identify facial videos, while preserving age, gender, race, pose, and expression of the original video. 
Specifically, we inpaint faces by a diffusion-based generative model guided by high-level attribute recognition and motion-aware expression transfer. 
We then animate de-identified faces by video-driven animation, which accepts the de-identified face and the original video as input.
Extensive experiments on the datasets VoxCeleb2, CelebV-HQ, and HDTF, which include diverse facial dynamics, demonstrate the effectiveness of AnonNET in obfuscating identity while retaining visual realism and temporal consistency. The code of AnonNet will be publicly released. 




\end{abstract}

\begin{figure}[ht]
  \centering
  \scriptsize
  \resizebox{\columnwidth}{!}{%
  \begin{tabular}{c c}
    \textbf{CelebA-HQ} & \textbf{LFW} \\
    
    \includegraphics[width=0.45\linewidth]{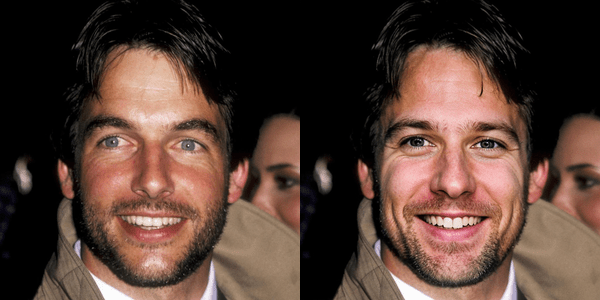} &
    \includegraphics[width=0.45\linewidth]{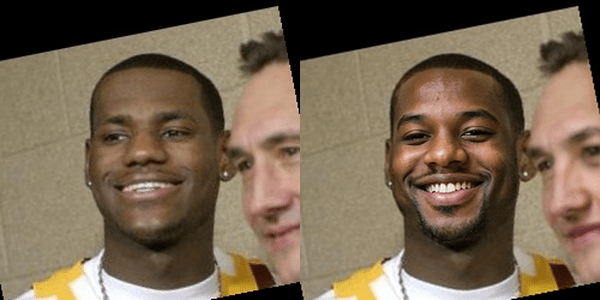} \\
    
    \includegraphics[width=0.45\linewidth]{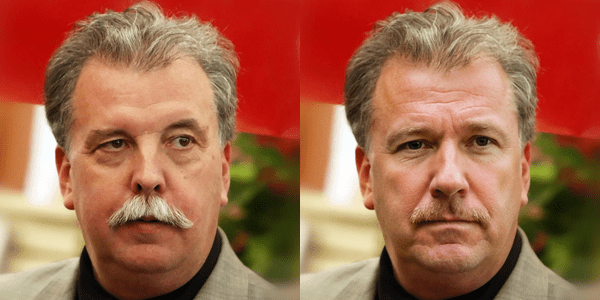} &
    \includegraphics[width=0.45\linewidth]{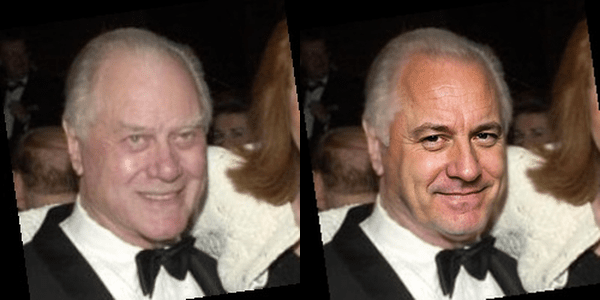} \\
    
    \includegraphics[width=0.45\linewidth]{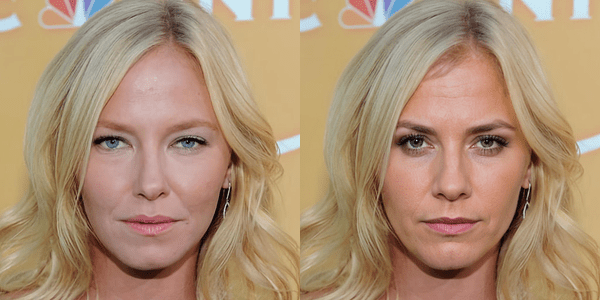} &
    \includegraphics[width=0.45\linewidth]{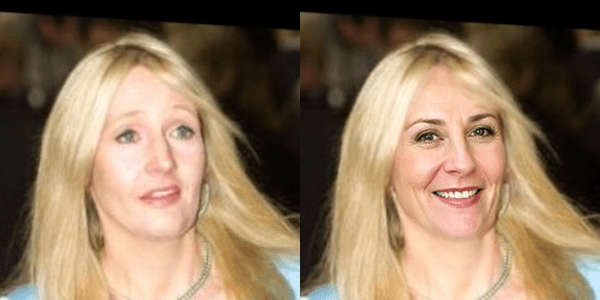} \\
    
    \includegraphics[width=0.45\linewidth]{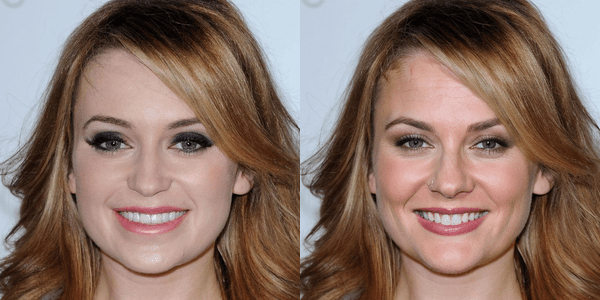} &
    \includegraphics[width=0.45\linewidth]{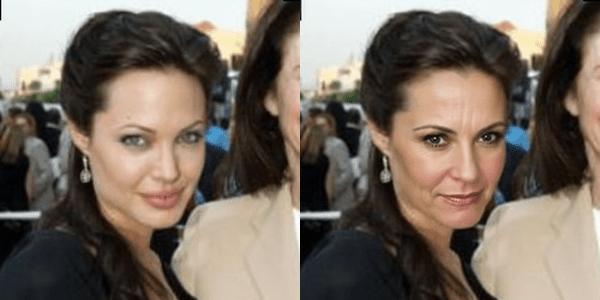} \\
  \end{tabular}
  }
  \caption{Qualitative comparison between original and anonymized face pairs. Left: CelebA-HQ. Right: LFW. Each row shows original/anonymized pairs.}
  \label{fig:celeba_lfw_duals}
\end{figure}

\section{Introduction}
Video anonymization is aimed at effectively obscuring identity-information, such as faces or voices, without compromising the integrity or usability of the content. Such anonymization has been fueled by ethical, legal or practically necessity - increasingly essential in a world, where facial images and videos have become omnipresent \cite{dantcheva:tel-03500318}.
For instance, medical therapy sessions recorded for research require video anonymization to protect patient identities, particularly facial features, while preserving related expressions and emotions, which are pertinent for research.


In addition, legal frameworks such as the General Data Protection Regulation (\textit{GDPR})\footnote{\url{https://gdpr-info.eu}} impose strict constraints on collection, processing, and dissemination of personal data, including biometric identifiers such as images and videos of the human face. More recently, the European Union's Artificial Intelligence Act (\textit{AI Act}))\footnote{\url{https://artificialintelligenceact.eu}}, adopted in 2024, introduced a tiered risk-based framework that places heightened scrutiny on AI systems handling biometric and identity-sensitive data. These evolving regulations reinforce the demand for anonymization methods that ensure privacy protection, while preserving the utility of data for downstream computer vision tasks. Traditional approaches including pixelation, blurring, and masking often degrade video quality and compromise associated applicability in such tasks~\cite{shoshitaishvili2015portrait}.

Modern deep generative models, especially Generative Adversarial Networks (GANs), have significantly advanced visual realism. \textit{Image anonymization} inpainting-based approaches, such as DeepPrivacy \cite{hukkelas2019deepprivacy} and DeepPrivacy2 \cite{hukkelas2023deepprivacy2} replace only sensitive facial regions, thereby preserving the surrounding content of the facial area. 
Conversely, fully synthetic pipelines such as FALCO~\cite{barattin2023attribute} generate artificial facial images, while preserving high-level attributes including age, gender and race, however may introduce inconsistencies in expression or struggle with robustness under diverse pose and lighting conditions due to reliance on GAN-inversion and matching in a synthetic latent space. 

W.r.t. \textit{video anonymization}, face-swapping methods \cite{zhang2020faceshifter} can inadvertently preserve identity-specific features, compromising unlinkability \cite{thies2019neural}. Finally, landmark-based motion transfer techniques risk motion artifacts, in cases when tracking is imperfect \cite{siarohin2019first}.

Motivated by the above, in this work we propose  \textit{AnonNET}, a multi-stage framework that (a) \textit{synthesizes new identities}, while preserving facial attributes. Our approach employs diffusion-based inpainting guided by structural priors for comprehensive identity obfuscation, avoiding the limitations of reference-based methods. Further, (b) a landmark-free motion transfer module ensures realistic expressions without relying on explicit keypoint tracking, thereby mitigating alignment fail cases. By restricting modifications to the face region, AnonNET retains original scene context, offering high-quality, privacy-preserving video anonymization. 

The main contributions of our work include the following.
\begin{itemize}
    \item A \textit{novel multi-stage framework} for video anonymization that synthesizes new facial identities while preserving age, gender, race, and expressions.
    \item A \textit{new dataset of anonymized videos} pertained to VoxCeleb, CelebV, and HDTF datasets, providing a valuable resource for future research in areas such as deepfake detection. 
    \item A \textit{comprehensive evaluation} of our pipeline against state-of-the-art models on \textit{image level}, as well as with regard to the re-identification, identity consistency, and expression-aware downstream utility on \textit{video level}. 
\end{itemize}

\section{Related Work}

\subsection{Image Anonymization}

\textbf{Traditional Techniques} include pixelation, blurring, as well as masking obscure facial features, in order to hinder identity recognition, preserving general image context. However, such techniques can significantly degrade images, impeding tasks such as expression analysis or attribute prediction \cite{newton2005preserving, kramer2019pixelation}. In addition, prior work has demonstrated that such anonymization techniques can be partially reversible \cite{mcpherson2016defeating}, raising concerns about related robustness in adversarial settings and the potential for identity leakage, rendering such methods insufficient for scenarios that demand both, privacy protection and downstream utility.

\textbf{Adversarial Techniques} employ \emph{adversarial training}, balancing image anonymization and utility preservation within a min-max framework. Nasr \textit{et al.}~\cite{nasr2018machine} used adversarial regularization to defend against membership inference, while Wu \textit{et al.}~\cite{wu2018towards, wu2019privacy} leveraged GANs to de-identify faces without compromising action recognition. Nonetheless, directly training GANs in the image space is challenging \textit{w.r.t.} fine-grained preservation of expressions and background details due to high pixel-space complexity.

\begin{figure*}[t]
  \centering
  \includegraphics[width=1\textwidth]{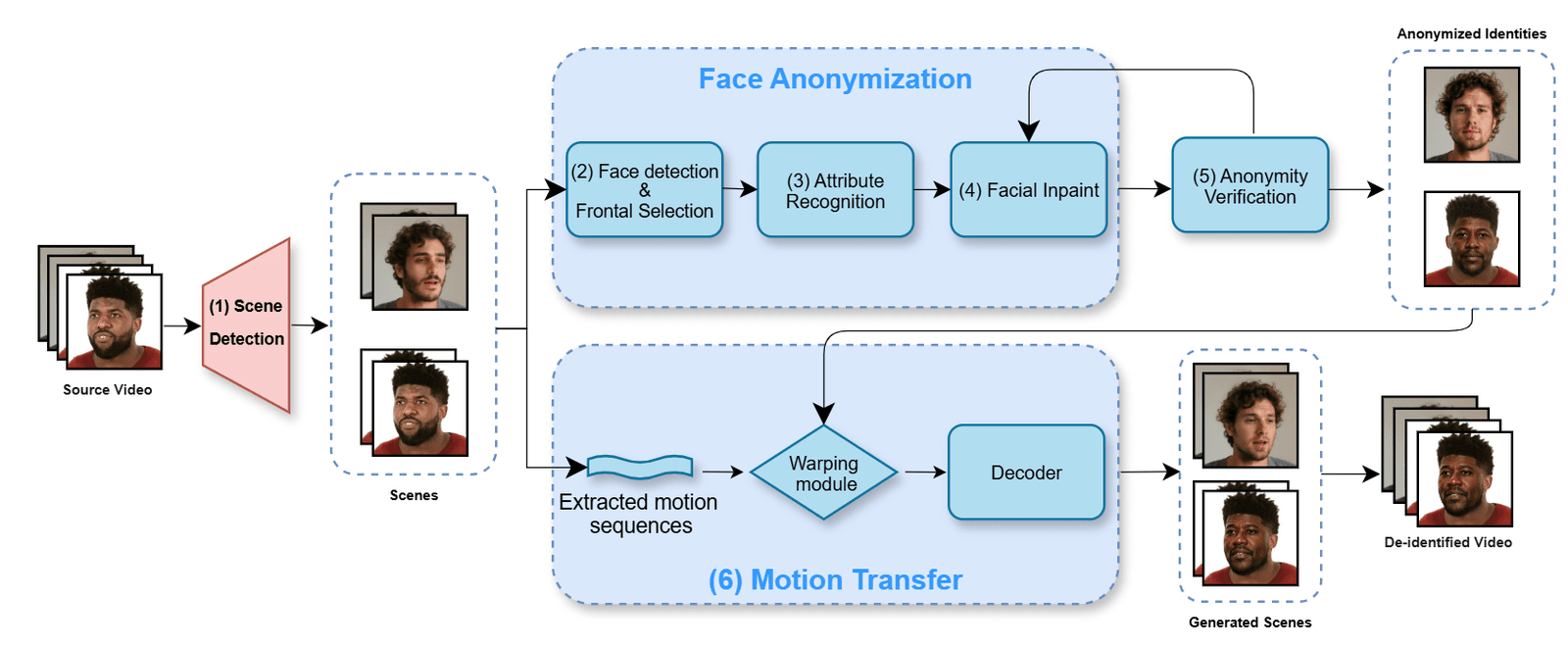}
  \caption{Overview of our multi-stage anonymization AnonNET-pipeline. (1) Scene changes are detected, and identities are tracked. (2) Faces are detected and single frontal frame is selected per scene identity. (3) Facial attributes are recognized. (4) A diffusion-based model inpaints the masked face. (5) Current anonymity is evaluated. (6) Landmark-free motion transfer reintroduces natural head movement. (7) Frames are reassembled for a coherent output video.}
  \label{fig:vid-anon}
\end{figure*}

To address these challenges, some methods manipulate latent representations \cite{le2022styleid}, disentangling identity-specific traits from other image-features. Nevertheless, such approaches may inadvertently preserve cues correlated with identity or require accurate facial landmarks or segmentation, which can be error-prone \cite{sun2018natural, hu2022protecting, maximov2020ciagan}. DeepPrivacy2 \cite{hukkelas2023deepprivacy2} extended a guided GAN framework, in order to anonymize full-body images, relying on precise facial segmentation and landmark detection. Failure cases in the latter can compromise both, anonymity and image quality.

\textbf{Diffusion-Based Approaches} iteratively refine noisy inputs, employing U-Net-like architectures to anonymize faces \cite{kung2024face, piano2024latent,Kung_2025_WACV}. Such methods often apply constrained transforms for obfuscation, yielding convincing results. Related limitations include the lack of control of diverse attributes such as age, gender, race, and expressions. Our approach similarly uses diffusion-based inpainting, however conditions it on user-specified attributes, granting fine-grained control over face appearance, while preserving non-identity aspects of the scene. Therefore, AnonNET enables flexible anonymization that balances privacy with visual fidelity and enables tasks such as emotion recognition or pose estimation. 

\subsection{Motion Transfer}
Motion transfer focuses on generating videos, with an appearance stemming from a reference image, guided by video-motion, \textit{e.g.,} head pose, expressions \cite{xu2024magicanimate, hu2024animate}. Early works used explicit keypoints or 3D models \cite{wang2022latent}, however experiencing tracking errors in complex motions. More recent approaches learn to warp or synthesize motion in the latent space, reducing reliance on landmark detection \cite{guo2024liveportrait}.

\textbf{Non-Diffusion} techniques such as FOMM \cite{siarohin2021motion}, MRAA \cite{zeng2023face}, SAFA \cite{wang2021safa}, or Face~vid2vid \cite{wang2021one} detect implicit keypoints to warp source images, however lack background stitching or fine eye/lip retargeting. They also are challenged by head poses. \\
Landmark-free approaches \cite{WANG_2020_WACV,Wang_2020_CVPR} aim at manipulating the latent space. 
For instance, the  \emph{Latent Image Animator} LIA \cite{10645735} builds a latent motion dictionary, whereas LivePortrait \cite{guo2024liveportrait} employs stitching and retargeting for flexible portrait animation. Such methods capture subtle expressions without the necessity of structural information. We note extreme motion or occlusion remain challenging.

\textbf{Additional Models} such as FADM~\cite{zeng2023face}, Face Adapter~\cite{han2024face}, AniPortrait~\cite{wei2024aniportrait}, X-Portrait~\cite{xie2024x}, and MegActor~\cite{yang2024megactor} achieve high-fidelity facial reenactment, often at the cost of increased computational complexity. These approaches frequently rely on 3D priors or explicit landmark extraction, rendering them less suitable for large-scale or real-time applications.

In contrast, AnonNET prioritizes computational efficiency, avoiding the necessity for 3D reconstruction or facial landmarks. This enables efficient anonymization, allowing for processing of longer video sequences. 

\subsection{Video Anonymization}
RID-TWIN \cite{mukherjee2024rid} used BLIP for face captioning and stable diffusion jointly with MediaPipe-based segment extraction. However, it focused on automatic de-identification rather than preserving specific attributes like age, gender, race, or expressions. SAFA was leveraged \cite{wang2021safa} for head motion transfer, which is prone to landmark-based errors.
SAFA used self-supervised landmark-like keypoints, however associated low resolution leads to motion and appearance artifacts.

\textbf{AI Stylization} \cite{yalccin2024empathy} constitutes a perceptual approach for anonymization, replacing facial realism with \textit{artistic} abstraction. After an initial facial feature randomization stage, the method applied cubist and painterly stylizations to anonymized faces, aiming to preserve emotional salience and enhance viewer empathy. However, the approach sacrificed photorealism entirely, as renderings tend to be stylized and visually inconsistent across frames. 

In contrast, our AnonNET-pipeline conditions diffusion-based inpainting on user-defined attribute priors, enabling consistent preservation of essential characteristics. Specifically, we adopt the \emph{landmark-free} motion transfer framework LIA and LivePortrait, in order to transfer expression and head pose. Additionally, we systematically detect \emph{scene changes} for improved temporal consistency and seamless anonymization across longer video sequences, therefore accommodating complex multi-scene videos effectively.

\section{Method}

\subsection{Overview of the Proposed Framework}
\label{sec:system_flow}

\begin{figure*}[t!]
  \centering
  \includegraphics[width=1\textwidth]
  {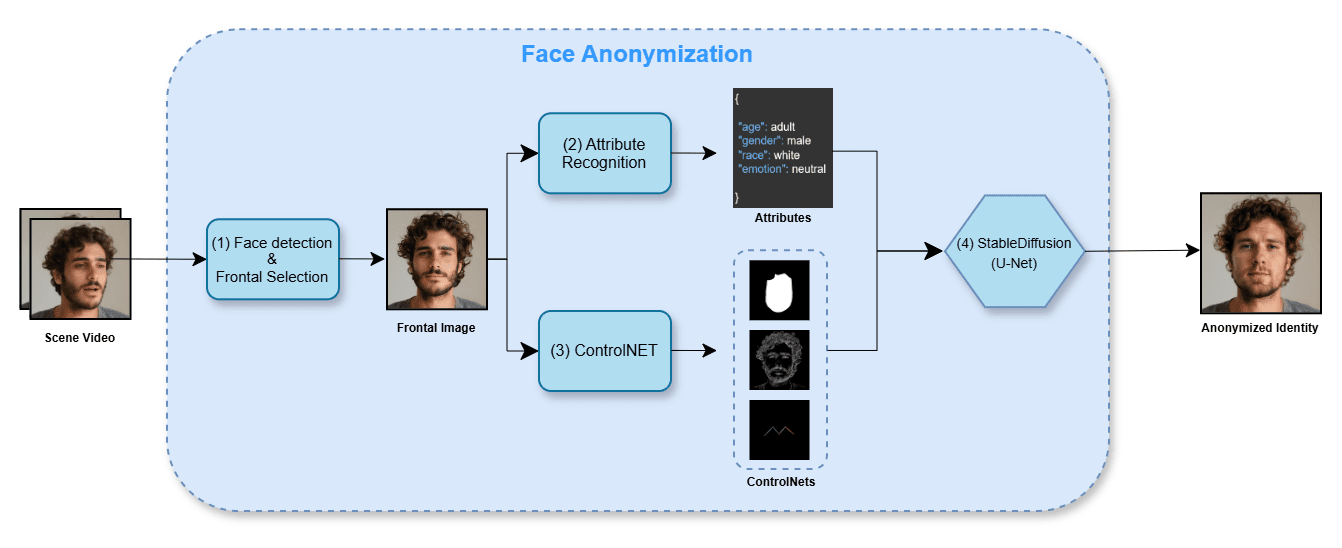}
\caption{Overview of the expression-consistent face anonymization module in AnonNET. (1) A face detection and frontal selection stage extracts a frame containing a frontal face from the input scene video. Then this frontal image is processed in parallel by two branches: (2) An attribute recognition module that infers semantic attributes such as age, gender, race, and expression; (3) a ControlNet module, which extracts structural guidance (\textit{e.g.,} face mask, lineart, pose) for conditioning the generative model. (4) Stable Diffusion based on U-Net synthesizes an anonymized face conditioned on both, extracted attributes and ControlNet features.}
  \label{fig:face-anon}
\end{figure*}

We propose a \textit{multi-stage} pipeline (see Figure~\ref{fig:vid-anon}), streamlined to obfuscate identity, while preserving attributes such as age, gender, race, and at the same time ensuring temporal consistency represented by expression and head poses. In particular, our AnonNET includes following stages.

\textbf{(1) Scene Detection \& Identity Clustering.}
We segment the input video via FFmpeg-based scene change detection, and then cluster faces across scenes using VGG-Face2 \cite{cao2018vggface2} embeddings and cosine-distance thresholds. Scenes containing the same individual share a consistent anonymized identity throughout the video.

\textbf{(2) Face Detection \& Frontal Selection.} RetinaFace \cite{deng2020retinaface} localizes the face region. A single representative frame, associated to a frontal pose, is selected per scene. This reduces flickering and computational burden by focusing anonymization on a single frame per segment.


\textbf{(3) Attribute Recognition.}  We estimate age, gender, race, and emotion via DeepFace \cite{serengil2024lightface}, retaining high-level features that do not reveal identity.


\textbf{(4) Diffusion-Based Inpainting.}
We adopt Realistic Vision V5.0 to inpaint the masked face guided by \textbf{ControlNets} for segmentation mask, lineart, and openpose maintain structural fidelity along with \textbf{Attribute-Conditioned Prompt} that we provide key attributes while discarding identity-specific details.
    

\textbf{(5) Anonymity Verification.}
To ensure identity obfuscation and avoid leakage at the end of the process, we include a verification module that verifies the cosine similarity between VGG-Face2 \cite{cao2018vggface2} embeddings of the original and anonymized images. In case that the similarity exceeds a threshold, the inpainting is re-triggered with higher stochasticity to enforce stronger anonymization.

\textbf{(6) Landmark-Free Motion Transfer.}
We select the frameworks \textbf{LIA} and \textbf{LivePortrait} to warp the anonymized face per frame, replicating natural head movements from the original video, without explicit landmark tracking. This approach mitigates flickering and alignment errors.

\textbf{(7) Video Reassembly.}
Processed frames are merged back into the original scene structure, retaining audio and background context. 

We note that resulting videos \textit{preserve facial attributes, exhibit strong identity obfuscation, and temporal coherence}, see Supplementary Material for videos.
Our modular design allows for each stage, namely scene detection, face detection, inpainting, motion transfer to be independently improved or replaced. We proceed to elaborate on each stage.

\subsection{Video Preprocessing and Scene Detection}
\label{sec:video_preprocessing}

\paragraph{(1) Scene Change Detection.}
We detect coarse scene boundaries using FFmpeg’s scene change filter (\texttt{select='gt(scene,X)'}), which flags transitions based on frame-wise histogram differences. To avoid over-segmentation, we refine these segments by computing mean RGB differences and merging visually similar intervals under a shared \texttt{scene\_id}. This simple yet effective two-stage strategy yields stable scene partitions and reduces redundancy, enabling consistent identity tracking and efficient anonymization across temporally coherent regions.



\paragraph{(2) Frontal Selection}
For each scene, we select a \emph{single} representative frame to serve as the anchor for anonymization and motion transfer. To ensure full facial coverage and minimize downstream hallucination, we prioritize frames with a frontal head pose, where both geometric structure and semantic attributes are fully visible.

We estimate head pose using the \texttt{face\_alignment} \cite{bulat2017far} library in 2D landmark mode, extracting 68 facial landmarks per frame. A subset of six key points (nose tip, chin, eye corners, mouth corners) is selected and matched to a predefined 3D face model. We then solve the perspective-n-point (PnP) problem via OpenCV's \texttt{solvePnP}, computing the 3D rotation vector of the head. Frames with fewer than 80\% of landmarks falling within image boundaries are discarded. Among valid candidates, the frame with the smallest absolute pitch and yaw is selected as the scene frontal frame.
This selection strategy ensures that identity obfuscation operates on a complete and unobstructed face. Since motion transfer is applied post-anonymization, any missing or occluded facial regions in the frontal frame would otherwise be synthesized without constraint—potentially leading to artifacts or semantic drift. 

\subsection{Face Detection and Attribute-Guided Prompt Generation}


\paragraph{(3) Attribute Recognition.} The localized face is then passed to the DeepFace library~\cite{serengil2024lightface}, in order to extract coarse demographic and affective attributes, including age, gender, race, and emotion. These attributes are used to guide the anonymization process in a non-identifying manner.

\subsection{Diffusion-Based Inpainting}
\label{sec:diffusion_inpaint}

\paragraph{(4) Diffusion-Based Inpainting.} Towards conditioning the diffusion-based inpainting, we construct a descriptive prompt, encoding the extracted attributes. Additionally, we apply a negative prompt to suppress undesired artifacts such as distortions, unrealistic textures, or cartoon-like features. The final prompt directs the generation toward a photorealistic, high-fidelity identity with preserved semantic traits. An example prompt is:
\begin{quote}
\texttt{A photorealistic portrait of a middle-aged Asian female, with a neutral expression.}
\end{quote}
We perform identity obfuscation via latent diffusion-based inpainting using Realistic Vision V5.0, a publicly available checkpoint based on Stable Diffusion v1.5~\cite{rombach2022high}.

ControlNet applies structural guidance from the head mask to confine synthesis to the face region while preserving surrounding content.

Lineart \& OpenPose ControlNets provide edge and pose priors to enforce geometric and expression fidelity during generation.

We use the DPMSolver++ scheduler for efficient denoising, typically over \texttt{20--70} steps with a guidance scale between \texttt{8--20}, depending on dataset characteristics. A VAE with perceptual reconstruction loss is used to map images between pixel and latent space, supporting visually coherent and detailed inpainting.

\subsection{Anonymity Verification and Motion Transfer} \label{sec:motion_transfer}
\paragraph{(5) Anonymity Verification.} To ensure successful identity obfuscation, we capture the cosine similarity between the VGG-Face2 embeddings of original and anonymized images. In case that the cosine distance score is below a threshold of $0.3$, indicating potential identity leakage, we re-trigger the inpainting process. In this second pass, we introduce greater stochasticity by increasing the prompt guidance scale and reducing ControlNet conditioning strength. Additionally, we extend the number of denoising steps by 5, in order to allow the diffusion process to deviate further from the original identity while still maintaining attribute consistency.

\paragraph{(6) Motion Transfer.} We combine landmark-free frameworks (\textbf{LIA}, \textbf{LivePortrait}) with scene stitching and eye/lip retargeting to animate anonymized faces across frames:
\begin{enumerate}
  \item \textbf{Encoding:} The original (non-anonymized) source frame is encoded to obtain a latent motion representation $z_{s \to r}$, capturing pose and expression dynamics.
  \item \textbf{Framewise Motion Codes:} For each target frame, motion offsets $\mathbf{w}_{r \to d}$ are predicted based on pose and expression changes.
  \item \textbf{Flow Field Synthesis:} The anonymized source frame image is encoded, and the learned flow map (from $z_{s \to r} + \mathbf{w}_{r \to d}$) is used to warp it, transferring motion to the anonymized face, while preserving appearance.
  \item \textbf{Refinement:} LivePortrait optionally enhances eye and lip dynamics (\textit{e.g.,} blinks, speech) for improved realism.
\end{enumerate}
By decoupling motion from identity and operating in latent and flow-guided spaces, these approaches avoid explicit landmark detection, reduce artifacts, and increase robustness to rapid motion and partial occlusion, while maintaining temporal consistency.


\paragraph{(5) Video Reassembly.} Finally, we reassemble processed scenes into the final video. Specifically, each anonymized segment is integrated using original timestamps, ensuring proper alignment.
We note that the audio is unaltered and resynchronized to preserve speech and background sounds.

Scenes with no faces remain untouched; multi-person scenarios are deferred for future work due to the complexities of simultaneous multi-face motion transfer.

\section{Results}
\subsection{Experimental Setup}

We proceed to comprehensively evaluate AnonNET's performance and compare related results to state-of-the-art anonymization methods, providing quantitative analysis on identity obfuscation 
and attribute retention. 
Further, an ablation study illustrates the impact of core components. 

\subsubsection{\textbf{Datasets}}

We evaluate our framework on following two widely used \textit{image} datasets.

 \textbf{CelebA-HQ}~\cite{karras2017progressive} contains 30,000 high-resolution face images annotated with 40 facial attributes, including age, gender, race, and appearance traits. The diversity in pose and lighting allow for our evaluation on attribute-preserving anonymization.
 
 \textbf{LFW}~\cite{huang2008labeled} includes 13,233 images of 5,749 identities captured in unconstrained conditions. We focus on identity obfuscation and generalization under varying image quality and occlusions.

For\textit{ video}-based anonymization, we additionally use following datasets.

 \textbf{CelebV-HQ}~\cite{zhu2022celebv} constitutes a curated high-resolution video face dataset. 
 
 \textbf{VoxCeleb2}~\cite{chung2018voxceleb2} represents a subset of 50,000 clips featuring over thousands of identities in varied conditions.
 
    \textbf{HDTF}~\cite{zhang2021flow} comprises expressive head motion and fine-grained lip sync.

\subsubsection{\textbf{Comparative Methods}}

We compare AnonNET against following \textit{image anonymization} frameworks.
   DeepPrivacy2~\cite{hukkelas2023deepprivacy2} is prominent for removing identity cues, while preserving contextual and structural consistency. At the same time CIAGAN~\cite{maximov2020ciagan} represents a competitive former approach that modifies latent identity features while retaining key facial attributes.

In addition, \textit{w.r.t.} \textit{video-anonymization}, we compare AnonNET to
RID-TWIN~\cite{mukherjee2024rid} that constitutes an end-to-end video anonymization approach with temporal coherence.

\subsubsection{\textbf{Implementation Details}}

AnonNET integrates a multi-stage pipeline with tailored configurations per component:

\textbf{Anonymization.} Based on Realistic Vision V5.0 with:
    \begin{itemize}
        \item Denoising steps: 20–70 (dataset-specific tuning)
        \item Guidance scale: 8–20 (for prompt expressiveness)
        \item ControlNets: Segmentation, LineArt, and OpenPose
        \item DPMSolver Scheduler for accelerated sampling
    \end{itemize}

\textbf{Motion Transfer.} Landmark-free video animation via LivePortrait~\cite{guo2024liveportrait} and LIA~\cite{wang2022latent}, enabling smooth expression preservation across frames.

\textbf{Computational Time.} Anonymizing the 50,000-clip VoxCeleb2 subset takes approximately 160 hours with LIA and 185 hours with LivePortrait, using a single A100 GPU.

\begin{figure}[b]
  \scriptsize
  \centering
  \resizebox{\columnwidth}{!}{%
  \begin{tabular}{c c c c c c}
    \multicolumn{6}{c}{\includegraphics[width=\linewidth]{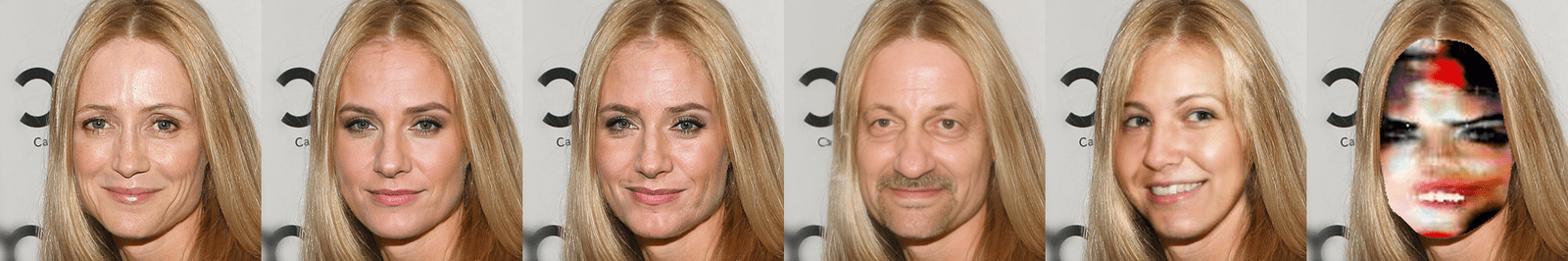}} \\
    \multicolumn{6}{c}{\includegraphics[width=\linewidth]{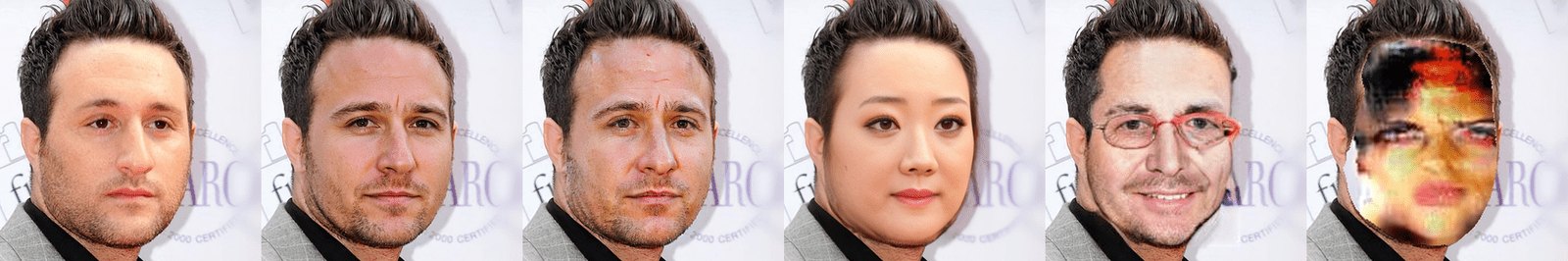}} \\
    \multicolumn{6}{c}{\includegraphics[width=\linewidth]{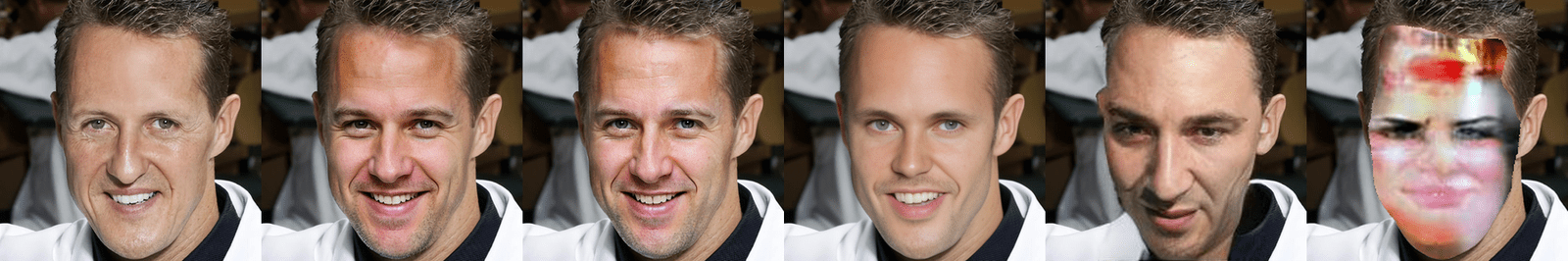}} \\
    \multicolumn{6}{c}{\includegraphics[width=\linewidth]{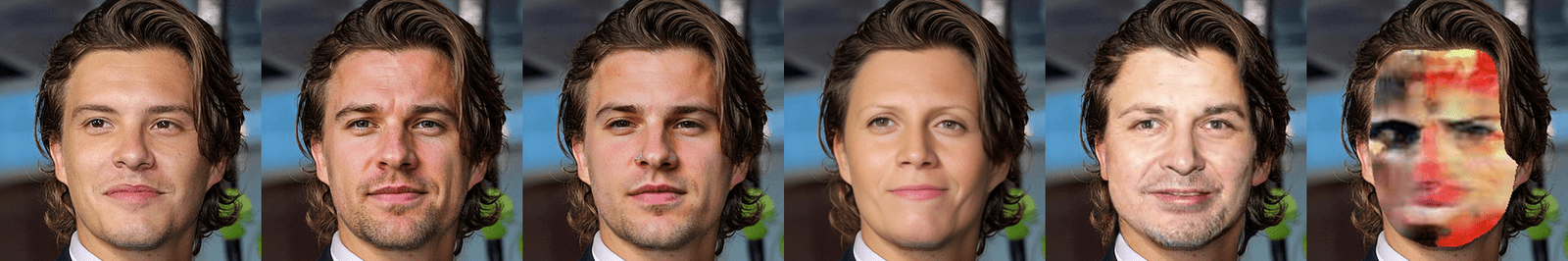}} \\
    \hspace{2.8mm}\textbf{Original} \hspace{1mm} & 
    \textbf{Ours \textit{35}} \hspace{1mm}& 
    \hspace{1mm}\textbf{Ours \textit{20}}\hspace{1mm} & 
    \textbf{FAMS}\hspace{4mm} & 
    \textbf{DP2} \hspace{1mm}& 
    \textbf{CIAGAN} \\
  \end{tabular}
  }
  \caption{Qualitative face anonymization results pertained to the CelebA-HQ dataset. Each row corresponds to an input image (left column), and columns show outputs from various image-anonymization methods.}
  \label{fig:comp_cele}
\end{figure}

\subsection{Evaluation Metrics}

We adopt an evaluation framework, aimed at assessing identity obfuscation, attribute retention, and visual quality.

 \textbf{Re-identification Rate (Re@1).} To evaluate identity leakage, we measure the rank-1 re-identification accuracy using face embeddings extracted from \textbf{VGGFace2}~\cite{cao2018vggface2} and \textbf{CASIA-WebFace}~\cite{yi2014learning} models. For each anonymized image, we compute its embedding and retrieve the closest match from the original image set based on cosine similarity. A sample is considered successfully re-identified if its nearest neighbor corresponds to the same identity as the original. The final Re@1 score is computed as the ratio of correctly re-identified samples over the total number of anonymized images. Lower scores indicate stronger identity obfuscation.

     \textbf{Image Quality and Aesthetics.} We use the \textbf{Q-Align/One-Align}~\cite{wu2023q} 
     metric to estimate perceptual quality (Qual) and visual appeal (Aes). These scores are averaged across all images and videos.

    \textbf{Pose and Gaze Preservation.} For each image, facial landmarks are first localized with MTCNN. Pose angles (pitch/yaw) are extracted via Dlib’s face pose estimator. Gaze direction is then evaluated using \textbf{L2CS-Net}~\cite{abdelrahman2022l2cs}, computing the mean absolute error (MAE) between original and anonymized outputs.

     \textbf{Expression Preservation.} Expression labels are predicted pre- and post-anonymization using the \textbf{DeepFace} library~\cite{serengil2024lightface}. Accuracy is defined as the fraction of samples, where the predominant expression label is retained.

     \textbf{Temporal Identity Consistency (Video).} For each anonymized video, we compute the average cosine distance of DINO~\cite{caron2021emerging} embeddings between consecutive frames, comparing with the same metric on original videos. This assesses intra-video consistency post-anonymization.


\begin{table}[ht]
\centering
\footnotesize  
\resizebox{\columnwidth}{!}{%
\begin{tabular}{l|cc|cc}
\multicolumn{1}{c|}{\textbf{Encoding}} & \multicolumn{2}{c|}{\textbf{CelebA-HQ}} & \multicolumn{2}{c}{\textbf{LFW}} \\
\cline{2-5}
 & VGG\ensuremath{\,\downarrow} & CASIA\ensuremath{\,\downarrow} & VGG\ensuremath{\,\downarrow} & CASIA\ensuremath{\,\downarrow} \\
\hline
DeepPrivacy2~\cite{hukkelas2023deepprivacy2}  & 0.008 & \textbf{0.008} & 0.023 & 0.017 \\
DeepPrivacy~\cite{hukkelas2019deepprivacy}    & 0.011 & 0.036         & 0.015 & 0.027 \\
CIAGAN~\cite{maximov2020ciagan}               & \textbf{0.004} & 0.022 & \textbf{0.009} & \textbf{0.002} \\
FALCO~\cite{barattin2023attribute}            & 0.017 & 0.028         & 0.016 & 0.021 \\
CAMOUFLaGE~\cite{piano2024latent}             & 0.096 & 0.100         & 0.102 & 0.116 \\
AnonNET\textit{(steps = 20)}                  & 0.073 & 0.031         & 0.056 & 0.039 \\
AnonNET\textit{(steps = 35)}                  & 0.041 & 0.017         & 0.042 & 0.027 \\
\end{tabular}%
}
\caption{Re-identification rate employing VGGFace2 and CASIA pertaining to the CelebA-HQ~\cite{karras2017progressive} and LFW~\cite{huang2008labeled} datasets. Lower scores denote a lower similarity between anonymized and original images and are therefore better.}
\label{tab:reid}
\end{table}

\subsection{Quantitative Evaluation}

\paragraph{Re-identification Performance.}
Table~\ref{tab:reid} reports rank-1 re-identification accuracy (Re@1) employing VGGFace2 and CASIA-WebFace embeddings on CelebA-HQ and LFW. While CIAGAN and DeepPrivacy2 achieve the lowest scores overall, AnonNET (35 steps) remains competitive, with Re@1 values of 0.041 (VGG) on CelebA-HQ and 0.042 on LFW. Compared to CIAGAN and DeepPrivacy, AnonNET provides a consistent drop in re-identification, while retaining attribute fidelity and performs favorably relative to recent diffusion-based anonymization baselines such as FALCO and CAMOUFLaGE.


\paragraph{Perceptual Quality and Aesthetic Appeal.}
As shown in Table~\ref{tab:quality}, AnonNET outperforms all baselines on Q-Align quality and aesthetic scores across both datasets. \textit{W.r.t.} CelebA-HQ, it achieves the highest quality (4.164) and aesthetics (3.332) scores, exceeding both, DeepPrivacy2 and ground truth. \textit{W.r.t.} LFW, AnonNET similarly leads with 2.887 (Qual) and 1.939 (Aes), indicating strong generalization to unconstrained, low-resolution data. In contrast, CIAGAN yields lower perceptual scores than both, AnonNET and DeepPrivacy2, consistent with degradation as noted in prior results. Beyond preserving visual realism, AnonNET systematically reduces artifacts and low-quality regions, owing to its attribute-guided diffusion design. These results are coherent and photorealistic outputs enhance compatibility with downstream tasks such as expression recognition or affective computing.

\begin{figure}[ht]
\centering
\scriptsize
\resizebox{\columnwidth}{1.5cm}{%
  \begin{tabular}{c c c c c c c}
    \raisebox{25pt}{\hspace{15pt}\LARGE\textbf{Original}} &
    \includegraphics[width=0.3\linewidth]{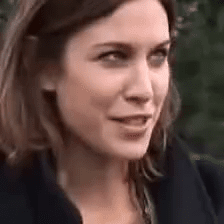} &
    \includegraphics[width=0.3\linewidth]{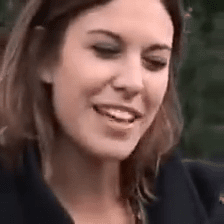} &
    \includegraphics[width=0.3\linewidth]{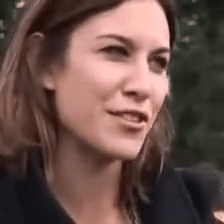} &
    \includegraphics[width=0.3\linewidth]{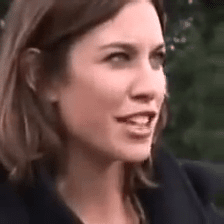} &
    \includegraphics[width=0.3\linewidth]{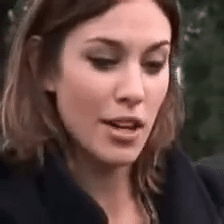} &
    \includegraphics[width=0.3\linewidth]{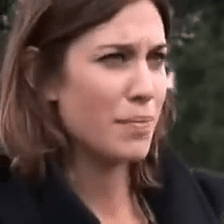} \\
    
    \raisebox{25pt}{\hspace{15pt}\LARGE\textbf{AnonNET}} &
    \includegraphics[width=0.3\linewidth]{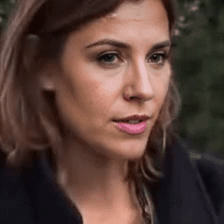} &
    \includegraphics[width=0.3\linewidth]{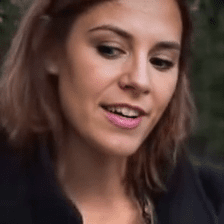} &
    \includegraphics[width=0.3\linewidth]{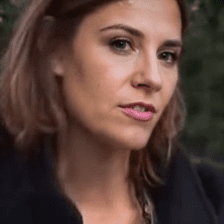} &
    \includegraphics[width=0.3\linewidth]{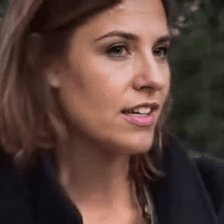} &
    \includegraphics[width=0.3\linewidth]{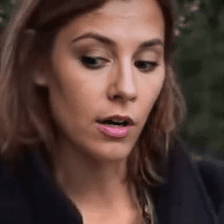} &
    \includegraphics[width=0.3\linewidth]{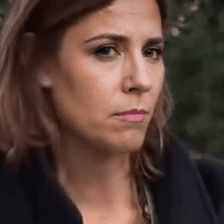} \\
  \end{tabular}%
}
\caption{Qualitative comparison of original and anonymized frames pertained to the VoxCeleb2 dataset using AnonNET. Each column shows original/anonymized pairs.}
\label{fig:vox_comp}
\end{figure}

\begin{table}[ht]
\centering
\begin{tabular}{l|cc|cc}
\multicolumn{1}{c|}{\textbf{Encoding}} & \multicolumn{2}{c|}{\textbf{CelebA-HQ}} & \multicolumn{2}{c}{\textbf{LFW}} \\
\cline{2-5}
 & Qual\ensuremath{\,\uparrow} & Aes\ensuremath{\,\uparrow} & Qual\ensuremath{\,\uparrow} & Aes\ensuremath{\,\uparrow} \\
\hline
GT                                           & 4.035 & 2.932    & 2.047 & 1.191    \\
DeepPrivacy2                                 & 3.551 & 1.875    & 2.025 & 1.099    \\
CIAGAN                                       & 1.011 & 1.361    & 1.006 & 1.466    \\
AnonNET\textit{(steps = 20)}                 & 4.074 & 3.055   & \textbf{2.914} & 1.904    \\
AnonNET\textit{(steps = 35)}                 & \textbf{4.164} & \textbf{3.332}    & 2.887 & \textbf{1.939}    \\
\end{tabular}
\caption{Quality and aesthetics scores for anonymized images of CelebA-HQ and LFW.}
\label{tab:quality}
\end{table}

\paragraph{Trade-off.}
The above results confirm that AnonNET offers a  privacy–utility trade-off: while not achieving the absolute lowest Re@1, it provides superior image quality. 

\paragraph{Video-level Evaluation.}
Table~\ref{tab:video} compares identity preservation, perceptual quality, and aesthetics between ground truth videos and those anonymized by AnonNET. Across all three datasets, AnonNET improves both, quality and aesthetics scores over the original videos while maintaining comparable levels of identity suppression. On CelebV-HQ and HDTF, our method achieves higher quality (\textit{e.g.,} 4.153 versus 4.045 on HDTF) and aesthetics (\textit{e.g.}, 3.021 versus 2.938), with only marginal differences in identity preservation.

\textit{W.r.t.} VoxCeleb2, which encompasses low resolution and challenging visual settings in related videos, AnonNET produces clean and coherently anonymized faces, raising quality from 2.493 to 2.859 and aesthetics from 1.606 to 1.949. These results highlight the robustness of our pipeline, even in unconstrained settings. Indeed, the new synthesized face improves structural integrity and overall visual consistency, rendering anonymized videos amenable to downstream tasks such as tracking or expression analysis.

\begin{table}[ht]
\centering
\resizebox{\columnwidth}{!}{%
\begin{tabular}{l|ccc|ccc}
\multicolumn{1}{c|}{\textbf{Dataset}} & \multicolumn{3}{c|}{\textbf{GT}} & \multicolumn{3}{c}{\textbf{AnonNET}} \\
\cline{2-7}
 & id\_pres\ensuremath{\,\downarrow} & qual\ensuremath{\,\uparrow} & aes\ensuremath{\,\uparrow} & id\_pres\ensuremath{\,\downarrow} & qual\ensuremath{\,\uparrow} & aes\ensuremath{\,\uparrow} \\
\hline
CelebV-HQ   & \textbf{0.011} & 3.800 & 2.718 & 0.013 & \textbf{3.907} & \textbf{2.886} \\
VoxCeleb2   & \textbf{0.021} & 2.493 & 1.606 & 0.022 & \textbf{2.859} & \textbf{1.949} \\
HDTF        & 0.008 & 4.045 & 2.938 & \textbf{0.007} & \textbf{4.153} & \textbf{3.021} \\
\end{tabular}%
}
\caption{Comparison of identity preservation, quality, and aesthetics in videos for ground truth and AnonNET across datasets. Lower is better for identity preservation (id\_pres); higher is better for quality (qual) and aesthetics (aes).}
\label{tab:video}
\end{table}

\begin{figure}[ht]
\centering
\scriptsize
\begin{tabular}{cc}
  \includegraphics[width=0.35\linewidth]{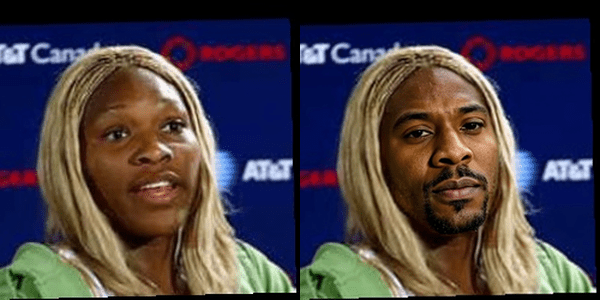} &
  \includegraphics[width=0.35\linewidth]{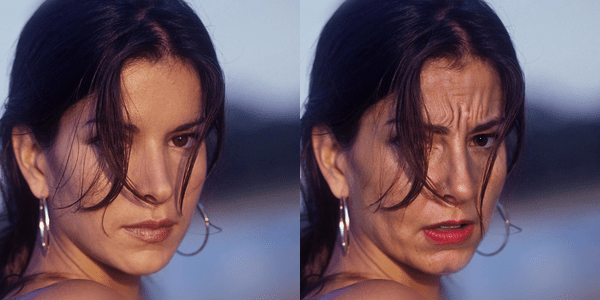} \\
  \textbf{(a)} Gender mismatch & \textbf{(b)} Expression mismatch \\[5pt]
  \multicolumn{2}{c}{
    \includegraphics[width=0.35\linewidth]{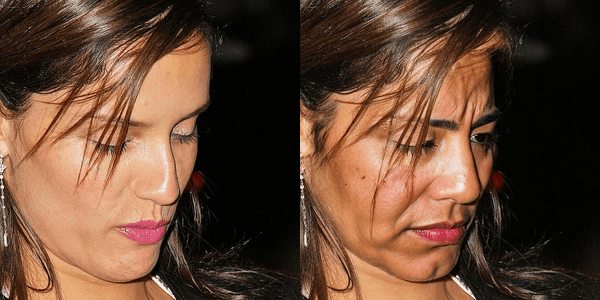}
  } \\
  \multicolumn{2}{c}{\textbf{(c)} Gaze inconsistency}
\end{tabular}
\caption{Comparison of original and anonymized images illustrating limitations of the proposed AnonNET framework.}
\label{fig:limitations}
\end{figure}

As shown in Figure~\ref{fig:vox_comp}, AnonNET reconstructs sharper facial features and preserves expressions more consistently than the original VoxCeleb2 frames, which suffer from heavy blur and compression. This visual improvement, especially in motion-rich regions, renders previously unusable videos viable for downstream tasks such as expression analysis or video reenactment.

Table~\ref{tab:pose_gaze} shows that AnonNET, guided by OpenPose, achieves superior pose preservation and competitive gaze alignment. Even with fewer denoising steps, it outperforms all baselines, highlighting its efficiency and motion consistency.


\begin{table}[ht]
\centering
\begin{tabular}{l|cc}
\multicolumn{1}{c|}{\textbf{Dataset}} & \multicolumn{2}{c}{\textbf{CelebA-HQ}} \\
\cline{2-3}
 & Pose\ensuremath{\,\downarrow}  & Gaze\ensuremath{\,\downarrow} \\
\hline
DeepPrivacy2                                    & 0.140 & 0.244 \\
FALCO                                           & 0.088 & 0.258 \\
FAMS\cite{Kung_2025_WACV}                                         & 0.048 & \textbf{0.161} \\
AnonNET\textit{(steps = 20)}                    & \textbf{0.014} & 0.187  \\
AnonNET\textit{(steps = 35)}                    & 0.015 & 0.172 \\
\end{tabular}
\caption{Pose and gaze preservation (lower is better) on CelebA-HQ.}
\label{tab:pose_gaze}
\end{table}

Table~\ref{tab:celeba_lfw_anonymization} summarizes AnonNET’s performance on CelebA-HQ and LFW. our proposed method achieves near-perfect anonymization rates with no detection failures. Attribute preservation remains high across both datasets, particularly for gender and race, while expression accuracy is lower on LFW due to its greater variability and resolution constraints. These results confirm AnonNET’s robustness across datasets with differing visual and demographic characteristics.

\begin{table}[ht]
\centering
\begin{tabular}{l|cc}
\multicolumn{1}{c|}{\textbf{Metric}} & \textbf{CelebA-HQ} & \textbf{LFW} \\
\hline
Total images              & 30,000  & 13,233 \\
Successfully anonymized   & 29,997  & 12,912 \\
Anonymization failures    & 3       & 321 \\
Face detection failures   & 0       & 0 \\
\hline
Race (\%)                 & 79.5    & 87.1 \\
Gender (\%)               & 99.4    & 99.3 \\
Age (mean $\pm$ std)      & (1.87, 4.23) & (2.69, 6.29) \\
Expression (\%)           & 74.7    & 52.9 \\
\end{tabular}
\caption{Anonymization statistics and attribute preservation accuracy on CelebA-HQ and LFW datasets.}
\label{tab:celeba_lfw_anonymization}
\end{table}

\paragraph{Limitations.} Figure~\ref{fig:limitations} highlights failure cases of AnonNET. Since attribute guidance relies on pretrained recognition networks, errors in gender, expression, or gaze estimation can propagate to the anonymized output. These limitations suggest the need for more robust or fine-tuned attribute predictors, especially for handling edge cases and underrepresented demographics.


\begin{figure}[ht]
\centering
\scriptsize
\resizebox{\columnwidth}{3.5cm}{%
\begin{tabular}{c}
  \includegraphics[width=0.3\linewidth]{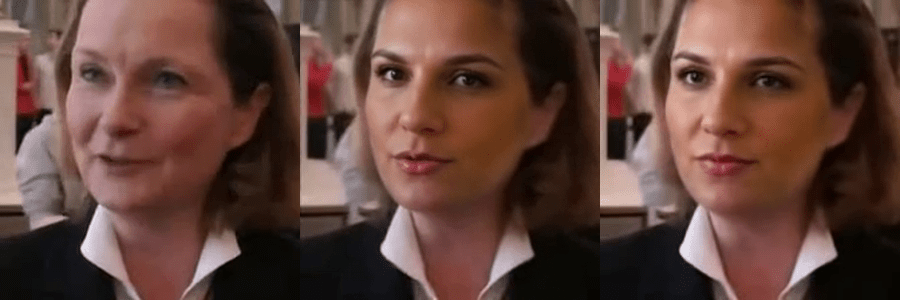} \\
  \includegraphics[width=0.3\linewidth]{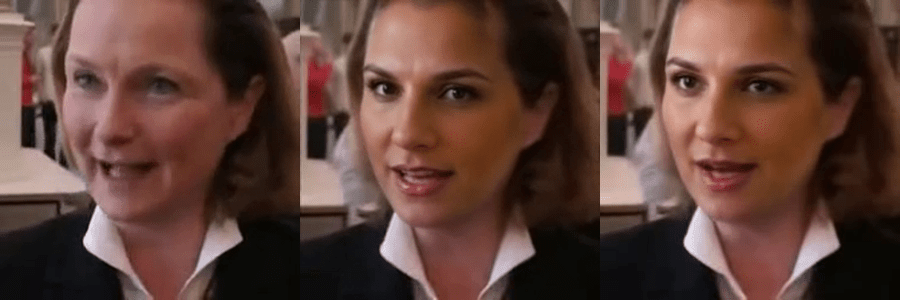} \\
  \includegraphics[width=0.3\linewidth]{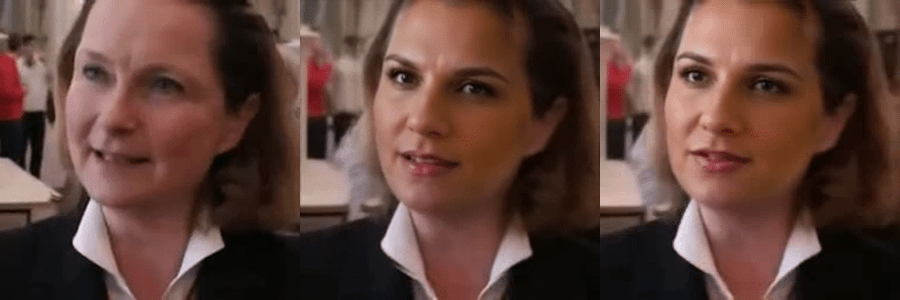} \\
\end{tabular}
}

\caption{Qualitative comparison of motion transfer models. Each row corresponds to following frames of a video, columns correspond Original, Live Portrait, and LIA, respectively.}
\label{fig:qualitative_sequences}
\end{figure}

\section{Conclusions}

In this work, we introduced AnonNET, a unified multi-stage framework for anonymizing talking head videos, placing emphasis on preserving key facial attributes. We presented extensive evaluations, demonstrating the ability of AnonNET to obfuscate identity, while allowing for further analysis. 
As opposed to the state of the art, AnonNET is robust to diverse poses, lighting conditions, and motion dynamics, rendering it suitable for real-world applications such as journalism, therapy, and human-computer interaction.


Future work will explore extending AnonNET to full-body video anonymization, incorporating audio-driven synchronization for improved lip consistency, and enhancing expression preservation in more dynamic conversational settings. These directions aim to further broaden the applicability of anonymized video content in sensitive or privacy-critical domains.



{
    \small
    \bibliographystyle{ieeenat_fullname}
    \bibliography{references}
}

\end{document}